\documentclass[a4paper, 10pt, conference]{ieeeconf}      

\IEEEoverridecommandlockouts                              

\usepackage{graphics} 
\usepackage{epsfig} 
\usepackage{mathptmx} 
\usepackage{times} 
\usepackage{amsmath} 
\usepackage{amssymb}  
\usepackage{subfigure}
\title{\LARGE \bf Longitudinal Dynamic versus Kinematic Models for Car-Following Control Using Deep Reinforcement Learning}

\author{Yuan Lin$^{1}$, John McPhee$^{2}$, and Nasser L. Azad$^{3}$
\thanks{$^{1}$Dr. Yuan Lin is a Postdoctoral Fellow in the Systems Design Engineering Department at University of Waterloo, Ontario, Canada N2L 3G1.
        {\tt\small y428lin@uwaterloo.ca}}%
\thanks{$^{2}$Dr. John McPhee is a Professor and Canada Research Chair in the Systems Design Engineering Department at University of Waterloo, Ontario, Canada N2L 3G1.
        {\tt\small mcphee@uwaterloo.ca}}%
\thanks{$^{3}$Dr. Nasser L. Azad is an Associate Professor in the Systems Design Engineering Department at University of Waterloo, Ontario, Canada N2L 3G1.
        {\tt\small nlashgarianazad@uwaterloo.ca}}%
}

\begin{document}

\maketitle
\thispagestyle{empty}
\pagestyle{empty}

\begin{abstract}

The majority of current studies on autonomous vehicle control via deep reinforcement learning (DRL) utilize point-mass kinematic models, neglecting vehicle dynamics which includes acceleration delay and acceleration command dynamics. The acceleration delay, which results from sensing and actuation delays, results in delayed execution of the control inputs. The acceleration command dynamics dictates that the actual vehicle acceleration does not rise up to the desired command acceleration instantaneously due to dynamics. In this work, we investigate the feasibility of applying DRL controllers trained using vehicle kinematic models to more realistic driving control with vehicle dynamics. We consider a particular longitudinal car-following control, i.e., Adaptive Cruise Control (ACC), problem solved via DRL using a point-mass kinematic model. When such a controller is applied to car following with vehicle dynamics, we observe significantly degraded car-following performance. Therefore, we redesign the DRL framework to accommodate the acceleration delay and acceleration command dynamics by adding the delayed control inputs and the actual vehicle acceleration to the reinforcement learning environment state, respectively. The training results show that the redesigned DRL controller results in near-optimal control performance of car following with vehicle dynamics considered when compared with dynamic programming solutions.

\end{abstract}

\section{INTRODUCTION}

Reinforcement learning is a goal-directed learning-based method that can be used for control tasks \cite{sutton2018reinforcement}. Reinforcement learning is formulated as a Markov Decision Process (MDP) wherein an agent takes an action based on the current environment state, and receives a reward as the environment moves to the next state due to the action taken. The goal of the reinforcement learning agent is to learn a state-action mapping policy that maximizes the long-term cumulative reward. DRL utilizes deep (multi-layer) neural nets to approximate the optimal state-action policy through trial and error as the agent interacts with the environment during training \cite{mnih2015human}. DRL has found recent breakthroughs as it surpassed humans in playing board games \cite{silver2016mastering}. DRL is actively evolving and various algorithms have been developed which include Deep Q Networks \cite{mnih2015human}, Deep Deterministic Policy Gradient (DDPG) \cite{lillicrap2015continuous}, Distributed Distributional Deterministic Policy Gradient \cite{barth2018distributed}, and Soft Actor Critic \cite{haarnoja2018soft}.

Connected and automated vehicles have become increasingly popular in academia and industry since DARPA urban challenge as autonomous driving could potentially become a reality \cite{urmson2008autonomous}. Fully autonomous driving is a challenging task since the transportation traffic can be dynamic, high-speed, and unpredictable. The Society of Automotive Engineers has defined multiple levels of automation as we progress from partial, such as Advanced Driver Assistance Systems (ADAS), to full automation. Current ADAS include ACC, lane-keeping assistance, lane-change assistance, emergency braking assistance, and driver drowsiness detection\cite{eskandarian2012}. Future highly automated vehicles shall be able to tackle more challenging traffic scenarios such as freeway on-ramp merging, intersection maneuver, and roundabout traversing.

Since DRL has been demonstrated to surpass humans in certain domains, it could potentially be suited to solve the challenging tasks in automated driving to achieve superhuman performance. Current literature has seen that DRL is used to tackle various traffic scenarios for automated driving. In \cite{wang2018autonomous}, Deep Q-learning is used to guide an autonomous vehicle to merge to freeway from on-ramp. In \cite{isele2018navigating,qiao2018automatically,qiao2018pomdp,li2018urban}, Deep Q Networks and/or DDPG allow an autonomous vehicle to maneuver through a single intersection while avoiding collisions. In \cite{wang2018reinforcement}, DRL is used to solve for the lane change maneuver. Other studies have also used DRL to train a single agent to handle a variety of driving tasks \cite{wolf2018adaptive,aradi2018policy}.

However, all the above-mentioned studies consider point-mass kinematic models of the vehicle, instead of vehicle dynamic models wherein acceleration delay and acceleration command dynamics are included. With acceleration delay, the reinforcement learning action such as the target acceleration is delayed in time; with acceleration command dynamics, the actual acceleration does not rise up to the target acceleration immediately \cite{jazar2017vehicle}. We acknowledge that acceleration command dynamics is being considered in a couple of most recent works that use DRL for vehicle control. In \cite{bucchel2018deep}, a longitudinal dynamic model is considered for predictive speed control using DDPG. In \cite{wei2018design}, a car-following controller is developed with acceleration command dynamics considered using DDPG by learning from naturalistic human-driving data. However, both studies did not investigate the impact of acceleration delay, which could degrade the control performance.

Regarding car-following control using DRL, there are other studies in the literature that have developed such controllers. In \cite{desjardins2011cooperative}, a cooperative car-following controller is developed using policy gradient with a single-hidden-layer neural net. In \cite{zhao2013supervised,zhu2018human}, human-like car-following controllers without considering vehicle dynamics are developed using deterministic policy gradient by learning from naturalistic human-driving data. To the best of our knowledge, there is currently no study that utilizes DRL to develop an ACC controller in simulation (not learning from naturalistic data).

There are studies in the literature that investigate delayed control inputs in non-deep reinforcement learning. It is suggested that the delay can negatively influence control performance if it is not considered in the reinforcement learning controller development \cite{schuitema2010design}. A few approaches have been proposed to cope with control delay for reinforcement learning. In \cite{hester2013texplore}, the environment state is augmented by adding the delayed control inputs, i.e., the actions in the delay interval which have not been executed, for developing a vehicle speed controller using reinforcement learning whose state-action mapping policy is a decision tree instead of a neural net. In \cite{walsh2009learning}, the authors proposed to learn the underlying dynamic system model so as to use the model to predict the future state after the delay for the purpose of determining the current control action. In \cite{schuitema2010control}, a memoryless method that exploits the delay length is proposed to directly learn the control action from the current environment state with the state-action mapping policy being a tile coding function instead of a neural net. There is currently no study that researches how a deep neural net trained in a no-control-delay environment responds to control delay. There is currently no work that develops a DRL controller with control delay considered.

The contribution of this work is studying the necessity and methodology of incorporating vehicle dynamics, which include both acceleration delay and acceleration command dynamics, in developing DRL controllers for automated vehicle control. We first investigate whether a DRL agent trained using vehicle kinematic models could be used for more realistic control with vehicle dynamics. We consider a particular car-following scenario wherein the preceding vehicle maintains a constant speed. As it shows that the DRL controller trained using a kinematic model causes significantly degraded performance when vehicle dynamics exists, we redesign the DRL controller by adding the delayed control inputs and the actual acceleration to the environment state \cite{katsikopoulos2003markov,hester2013texplore} to accommodate for vehicle dynamics.

\section{CAR-FOLLOWING PROBLEM FORMULATION}

In this section, we derive the state-space equations of the car-following control system so as to (1) understand how it could fit into the reinforcement learning framework with state-action mapping, and (2) use dynamic programming (DP) to compute the global optimal solutions for comparison with DRL solutions. DP is based on the state-space equations and checks all permissible state values to search for the global minimum cost for the control system \cite{naidu2002optimal}.

We acknowledge that the relatively easy car-following control problem may preferably be solved using classical control method instead of DRL which is more capable to solve more challenging control tasks such as freeway on-ramp merging. We choose the car-following control problem here because it can be explicitly modeled to obtain the state-space equations with which we can use DP to solve for the guaranteed global optimal solutions for comparison purposes. The DP solutions are critical because they serve as benchmarks with which we can evaluate the DRL controllers trained with either the vehicle dynamic or kinematic model. The other autonomous driving control tasks such as freeway on-ramp merging may not be explicitly modeled since they involve highly complex multi-vehicle interactions.

\begin{figure}[htbp]
\centering
\includegraphics[width=3.4in]{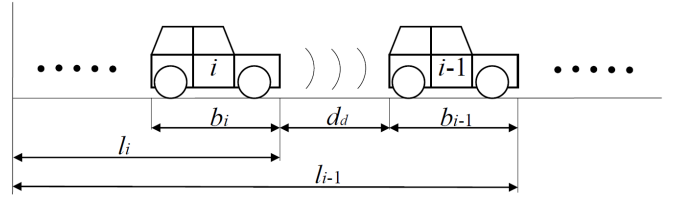}
\caption{Schematic for car following with a constant distance headway.}
\label{fig:schematic}
\end{figure}

We consider a simple car-following control problem wherein a following vehicle $i$ desires to maintain a constant distance headway $d_d$ between itself and its preceding vehicle $i-1$, see Fig.~\ref{fig:schematic}. The gap-keeping error dynamic equations of the car-following control system can be derived as:
\begin{align}
\begin{split}
e &= l_{i-1} - l_{i} - b_{i-1} - d_d\\
\dot{e} &= v_{i-1} - v_i\\
\ddot{e} &= a_{i-1} - a_i
\end{split}
\label{eq:e}
\end{align}
where $e$ is the error between the actual inter-vehicle distance and the desired distance headway $d_d$, $b_{i-1}$ is the vehicle body length of the preceding vehicle $i-1$, $l_{i-1}$ and $l_i$ are the distances traveled by the preceding and the following vehicles, respectively, $v_{i-1}$ and $v_i$ are the velocities of the preceding and following vehicles, respectively, $a_{i-1}$ and $a_i$ are the actual accelerations of the preceding and the following vehicles, respectively. For the state space representation, we define $x_1=e, x_2=\dot e$. Then
\begin{align}
\begin{split}
\dot{x}_1 &= x_2\\
\dot{x}_2 &= a_{i-1} - a_i
\end{split}
\label{eq:ss_e}
\end{align}

Assuming no vehicle-to-vehicle communication, the preceding vehicle's acceleration $a_{i-1}$ is unknown to the following vehicle. As the DRL algorithm used here is Deep Deterministic Policy Gradient which demands the system to be deterministic, we only consider preceding vehicle's speed to be a constant with $a_{i-1} = 0$. In fact, without knowing the preceding vehicle's acceleration, the system is not closed and the exact optimal solution could not be found. We found that even though the DRL neural nets are trained for this scenario in which the preceding vehicle has a constant speed, the trained neural nets could be applied to scenarios when the preceding vehicle accelerates or decelerates with acceptable gap-keeping errors. Since the purpose of this paper is to compare the use of dynamics versus kinematic models for vehicle control, we do not show such results here.

Now we consider using the vehicle kinematic and dynamic models for the control. For a point-mass kinematic model, the following vehicle's control input $u_i$ is exactly the acceleration $a_i$, i.e., $u_i = a_i$. The vehicle integrates and double-integrates over the control input (acceleration) for velocity and position updates, respectively. Thus, the state space representation when using a point-mass kinematic model is
\begin{align}
\begin{split}
\dot{x}_1 & = x_2\\
\dot{x}_2 &= -u_i
\end{split}
\label{eq:ss_km}
\end{align}

For a vehicle dynamic model, we adopt a simplified first-order system for the acceleration command dynamics from the current literature used for Toyota Prius and Volvo S60 \cite{ploeg2011design,lidstrom2012modular}, which is shown in Laplace Domain as
\begin{equation}
\frac{A_i(s)}{U_i(s)} = \frac{1}{\tau s + 1} e^{-\phi s}
\label{eq:a_L}
\end{equation}
where $s$ is the Laplace Transform variable, $A_i(s)$ and $U_i(s)$ are the Laplace Transforms of $a_i$ and $u_i$, respectively, $\tau$ is the time constant of the first-order system, and $\phi$ is the acceleration time delay. In time domain, the first-order system can be interpreted as
\begin{equation}
\dot{a}_i = \frac{u_i(-\phi) - a_i}{\tau}
\label{eq:a_t}
\end{equation}
where $u_i(-\phi)$ denotes that $u_i$ is delayed by $\phi$ in time. Introducing another state variable $x_3 = a_i$, the state space representation when using the dynamic model is
\begin{align}
\begin{split}
\dot{x}_1 &= x_2\\
\dot{x}_2 &= -x_3\\
\dot{x}_3 &= \frac{u_i(-\phi) - x_3}{\tau}
\end{split}
\label{eq:ss_dy}
\end{align}


The control goal is to minimize both the error and control effort, which is a common goal of classical control methods such as Linear Quadratic Regulator and Model Predictive Control. Here we define the absolute-value cost for the car-following control system as
\begin{equation}
J = \int_{0}^{t_f} (\alpha \frac{|e|}{e_{nmax}} + \beta \frac{|u_i|}{u_{max}}) dt
\label{eq:ss_cost}
\end{equation}
where $t_f$ is the terminal time, $|e|$ and $|u_i|$ denote the absolute values of the error and control input, respectively, $u_{max}$ is the allowed maximum of $|u_i|$, $e_{nmax}$ is the nominal maximum of $|e|$, and $\alpha$ and $\beta$ are coefficients that satisfy $\alpha>0$, $\beta>0$, and $\alpha+\beta=1$. The $\alpha$ and $\beta$ values can be adjusted so as to decide the weighting of minimizing the error over the control action in the combined cost. The $e_{nmax}$ is a nominal maximum because the gap-keeping error can be very large, especially during DRL training wherein the vehicle can have any acceleration behavior before it gets well trained, see the next section. We choose a sufficiently large $e_{nmax}$ to represent a maximum gap-keeping error of a general car-following transient state.

As both dynamic programming and reinforcement learning are based on discrete time, the above continuous-time equations are discretized using a forward Euler integrator. Note that the absolute-value cost is different than the quadratic cost for LQR and MPC. This is because, for DRL, absolute-value rewards lead to lower steady-state errors \cite{engel2014line}. As we want to compare DRL solutions with DP ones, the DP cost function needs to be the same as the DRL's.

\section{DEEP REINFORCEMENT LEARNING ALGORITHM}


In this section, we introduce the reinforcement learning framework and the specific DRL algorithm, DDPG (Deep Deterministic Policy Gradient), that we use to solve the above car-following control problem.

\subsection{Reinforcement Learning}

As stated in \cite{sutton2018reinforcement}, reinforcement learning is learning what to do, i.e., how to map states to actions, so as to maximize a numerical cumulative reward. The formulation of reinforcement learning is a Markov Decision Process. At each time step $t$, $t=0,1,2,...,T$, a reinforcement learning agent receives the environment state $s_t$, and on that basis selects an action $a_t$. As a consequence of the action, the agent receives a numerical reward $r(s_t,a_t)$ and finds itself in a new state $s_{t+1}$. In reinforcement learning, there are probability distributions for transitioning from a state to an action and for the corresponding reward, which are not illustrated here. The goal in reinforcement learning is to learn an optimal state-action mapping policy $\pi^\star$ that maximizes the expected cumulative discounted reward $R = E[\sum_{t=0}^{t=T} \gamma^{\,t} r(s_t,a_t)]$ with $E$ denoting the expectation of the probabilities. The symbol $\star$ denotes optimality. The Q-value, i.e., the state-action value, for time step $t$ is defined as the expected cumulative discounted reward calculated from time $t$, i.e., $Q(s_t,a_t) = E[\sum_{t}^{t=T} \gamma^{\,t} r(s_t,a_t)]$. Reinforcement learning problem is solved using Bellman's principle of optimality. That is, if the optimal state-action value for the next time step is known $Q^\star(s_{t+1},a_{t+1})$, then the optimal state-action value for the current time step can be solved by taking the action that maximizes $r(s_t,a_t) + Q^\star(s_{t+1},a_{t+1})$.

The reinforcement learning framework for the car-following control system is based on the state-space equations described in the previous section. The action of the reinforcement learning framework is the control input of the car-following control system $u_{i,t}$ for time $t$. The reward function is the negative value of the discretized absolute-value cost defined in Equation~\ref{eq:ss_cost} of the previous section.
\begin{equation}
\begin{split}
r(s_t,a_t) = - \alpha \frac{|e_{t+1}|}{e_{nmax}} - \beta \frac{|u_{i,t}|}{u_{max}}
\end{split}
\label{eq:drl_cost}
\end{equation}
With this expression, the reward value range is (-$\inf$,0]. We clip the reward to be in the range [-1,0] to avoid huge bumps in the gradient update of the policy and Q-value neural networks of DDPG. The huge bumps in the gradient update lead to training instability \cite{van2016learning}.


We consider 4 cases of the reinforcement learning framework as this work compares using dynamic versus kinematic models for autonomous vehicle control. For case 1, a kinematic model is used. Based on Equation~\ref{eq:ss_km}, only the gap-keeping error and error rate are sufficient to solve for the dynamic system. So the environment state vector is $s_t = [e_t, \dot{e}_t]$ for time step $t$.

For case 2, only acceleration delay is considered with no acceleration command dynamics. We consider this intermediate case for comparison purposes as well. In fact, for hybrid electric vehicles such as Toyota Prius \cite{ploeg2011design}, the time constant in the acceleration dynamics equation is small $\tau=0.1$s, which means that the vehicle responds to a desired acceleration very quickly. Also, for pure electric vehicles, the response is even faster. For such vehicles, the acceleration command dynamics results in little degradation in the DRL control performance, as we observed in our simulations. Therefore, case 2 may represent DRL control for hybrid and pure electric vehicles. For this case, we define the state vector as $s_t = [e_t, \dot{e}_t, u_{i,t-k},...u_{i,t-1}]$ with $k$ being the largest integer such that $k*\Delta t \leq \phi$ with $\Delta t = 0.1$s being one time step value. This means that we feed into the DRL agent the past control inputs that haven't been executed by the control system due to time delay. We expect the DRL agent to use these delayed control inputs to solve for the corresponding system responses that would happen in the future and predict the next optimal control input $u_{i,t}$.

For case 3, only acceleration command dynamics is considered with no acceleration delay $\phi = 0$. For this case, the time constant is $\tau=0.5$s, which applies to gas-engine vehicles such as Volvo S60 \cite{lidstrom2012modular}. We consider this intermediate case for comparison purposes. According to Equation~\ref{eq:ss_dy}, the state vector is $s_t = [e_t, \dot{e}_t, a_{i,t}]$ which includes the error, error rate, and the actual acceleration of the following vehicle.

\begin{table}[h]
\caption{Car-following control system parameter values.}
\begin{center}
\begin{tabular}{|c|c|}
\hline
Discrete time step $\Delta t$ & 0.1s\\
\hline
Nominal max error $e_{nmax}$ & 10m\\
\hline
Max control input $u_{max}$ & 2.6 m/s$^2$\\
\hline
Acceleration delay $\phi$ & 0.2s\\
\hline
Acceleration command dynamics time constant $\tau$ & 0.5s\\
\hline
Preceding vehicle constant speed $v_{i-1}$ & 30m/s\\
\hline
Following vehicle initial speed $v_i(t=0)$ & 27.5m/s\\
\hline
Initial gap-keeping error $e(t=0)$ & 2.5m\\
\hline
\end{tabular}
\end{center}
\label{table:para}
\end{table}

For case 4, both acceleration command dynamics and delay are considered. For this case, the time constant is also $\tau=0.5$s for gas-engine vehicles. The state vector is $s_t = [e_t, \dot{e}_t, a_{i,t}, u_{i,t-k},...u_{i,t-1}]$. Table~\ref{table:para} shows the parameter values for the car-following control system.

\subsection{Deep Deterministic Policy Gradient}

The DRL algorithm that we use is DDPG, which is exactly the same as proposed in \cite{lillicrap2015continuous}. Here we provide a brief description of the DDPG algorithm and we encourage the readers to read the original paper. The DDPG algorithm utilizes two deep neural networks: actor and critic networks. The actor network is for the state-action mapping policy $\mu(s_t|\theta^\pi)$ where $\theta^\pi$ denotes the actor neural net weight parameters, and the critic network is for Q-value function (cumulative discounted reward) $Q(s_t,a_t|\theta^Q)$ where $\theta^Q$ denotes the critic neural net weight parameters. DDPG concurrently learns the policy and Q-value function. For learning the Q-value (Q-learning), the Bellman's principle of optimality is followed to minimize the root-mean-squared loss $L_t = r(s_t,a_t) + Q(s_{t+1},\mu(s_{t+1}|\theta^\pi)) - Q(s_t,a_t|\theta^Q)$ using gradient descent. For learning the policy, gradient ascent is performed with respect to only the policy parameters $\theta^\pi$ to maximize the Q-value $Q(s_t,\mu(s_t|\theta^\pi))$.

\begin{table}[h]
\caption{Deep Deterministic Policy Gradient parameter values.}
\begin{center}
\begin{tabular}{|c|c|}
\hline
Target network update coefficient & 0.001\\
\hline
Reward discount factor & 0.99\\
\hline
Actor learning rate & 0.0001\\
\hline
Critic learning rate & 0.001\\
\hline
Experience replay memory size & 500000\\
\hline
Mini-batch size & 64\\
\hline
Actor Gaussian noise mean & 0\\
\hline
Actor Gaussian noise standard deviation & 0.02\\
\hline
\end{tabular}
\end{center}
\label{table:ddpg_para}
\end{table}

Target networks are adopted to stabilize training \cite{mnih2015human}. We use Gaussian noise for action exploration \cite{bucchel2018deep}. Mini-batch gradient descent is used \cite{lillicrap2015continuous}. Experience replay is used for stability concerns \cite{mnih2015human}. Batch normalization is used to accelerate learning by reducing internal covariant shift \cite{ioffe2015batch}. Please see Table~\ref{table:ddpg_para} for the DDPG algorithm parameter values.

\begin{figure}[htbp]
\centering
\includegraphics[width=3.4in]{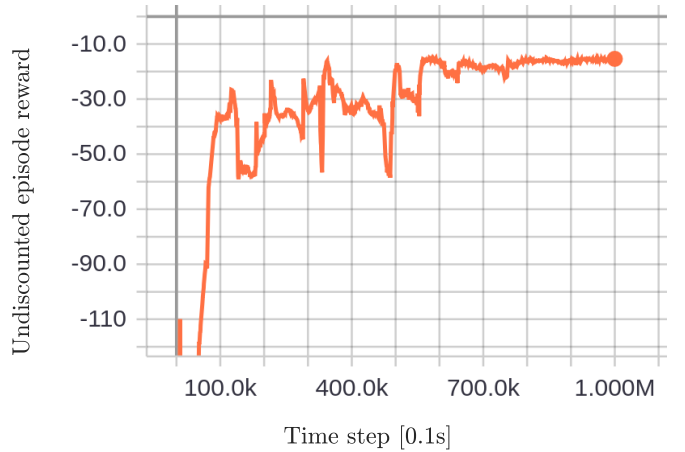}
\caption{Undiscounted episode reward for training with the vehicle kinematic model.}
\label{fig:reward}
\end{figure}

\begin{figure*}[htbp]
\centering
\includegraphics[width=7in]{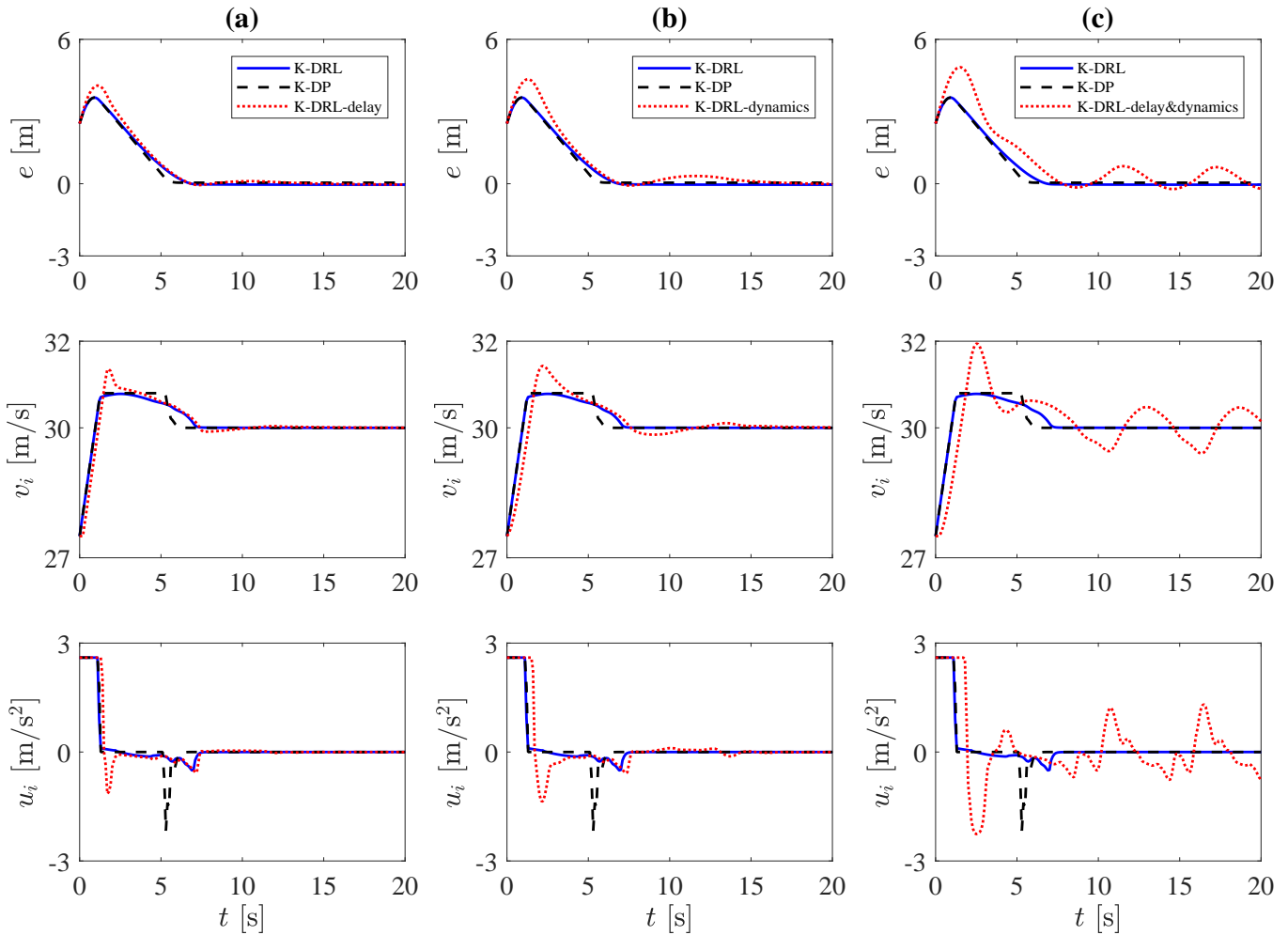}
\caption{Training and testing results for the DRL controller trained using the point-mass kinematic model (K-DRL). Columns (a), (b), and (c) show the results of testing this DRL controller for car-following control with just acceleration delay (K-DRL-delay), just acceleration command dynamics (K-DRL-dynamics), and both acceleration delay and command dynamics (K-DRL-delay\&dynamics), respectively. The variable $e$ is the gap-keeping error, $v_i$ is the following vehicle's velocity, and $u_i$ is the control input to the following vehicle. Note that the control input is equal to the actual acceleration for the kinematic model case. The preceding vehicle's constant speed is $v_{i-1}$=30m/s which is not shown here.}
\label{fig:K}
\end{figure*}

Both the actor and critic networks are neural nets with 2 hidden layers for all cases. For training with vehicle kinematics (case 1) and just acceleration command dynamics (case 3), the neural nets have 64 neurons for each hidden layer and the training time is 1 million time steps. For training with control delays (cases 2 and 4), the neural nets have 128 neurons for each hidden layer and the training time is 1.5 million time steps. For all cases, the training converges. Fig.~\ref{fig:reward} shows the undiscounted episode reward for case 1. The plots of the undiscounted episode rewards for all the other cases look similar to that for case 1, and are not shown here. We use the undiscounted episode reward since it allows us to track changes for the latter part of the car-following errors easily. Note that, with the discount factor, the last reward at 20 seconds (200 time steps) of one episode is discounted by $0.99^{200} = 0.134$. We use OpenAI Gym \cite{brockman2016openai} for creating the training environment and Stable Baselines \cite{stable-baselines} for the DDPG training.

\begin{figure*}[htbp]
\centering
\includegraphics[width=7in]{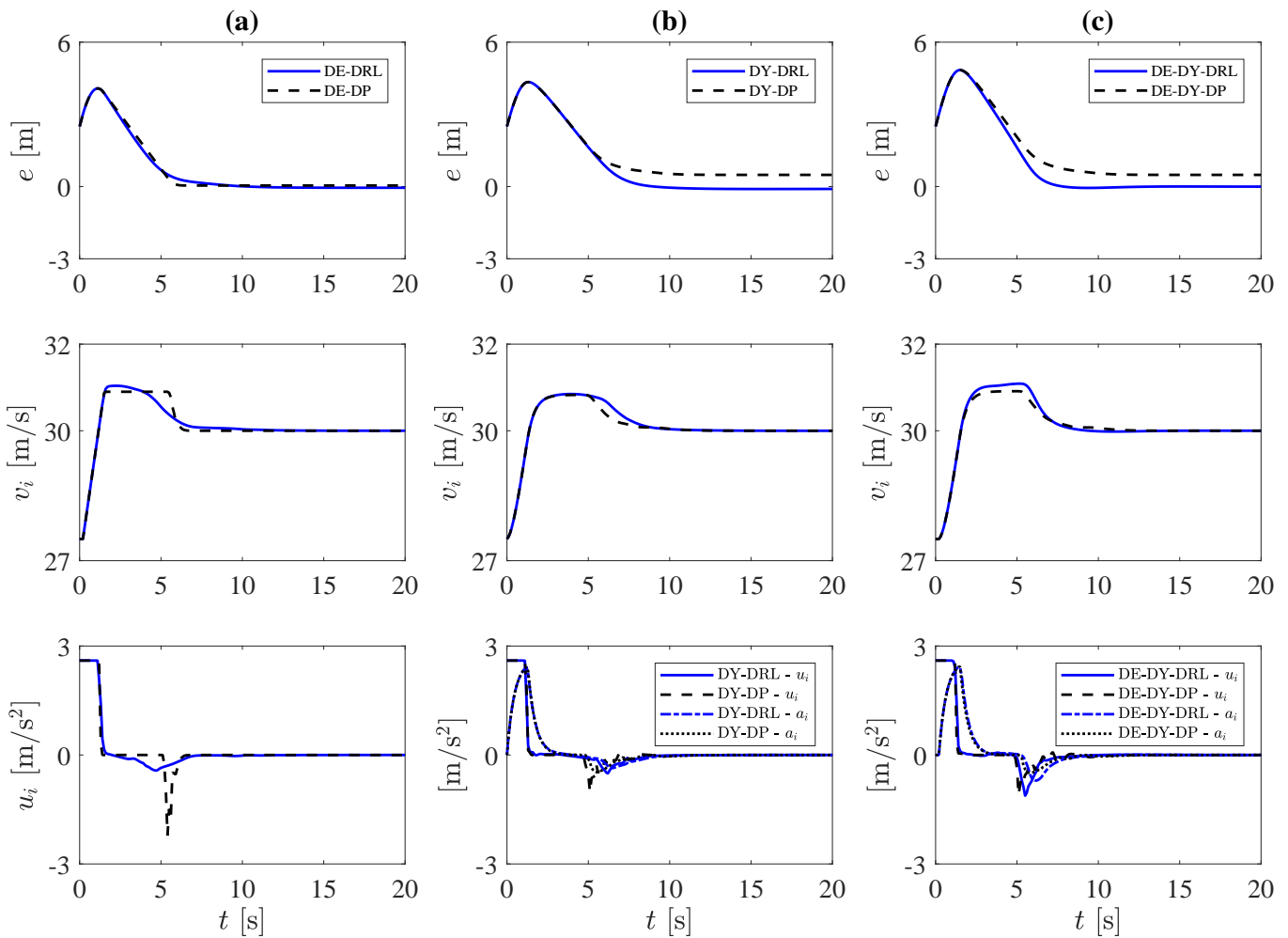}
\caption{DRL controller results when trained with just acceleration delay (DE-DRL, column (a)), just acceleration command dynamics (DY-DRL, column (b)), and both acceleration delay and command dynamics (DE-DY-DRL, column (c)). The variable $e$ is the gap-keeping error, $v_i$ is the following vehicle's velocity, $u_i$ is the control input to the following vehicle, and $a_i$ is the actual acceleration of the following vehicle. The preceding vehicle's constant speed is $v_{i-1}$=30m/s which is not shown here.}
\label{fig:D}
\end{figure*}

\section{RESULTS}

In this section, the DRL results for the above mentioned 4 cases are presented. We also present the DP results which are the global optimal solutions for all cases for comparison purposes. We first present DRL and DP results for the car-following control with a point-mass kinematic model and the results of applying this kinematics-model-trained DRL controller to car-following control with vehicle dynamics, which are shown in Fig.~\ref{fig:K}. We then present the results of our proposed solution to deal with acceleration delay and acceleration command dynamics by adding the delayed control inputs and the current actual acceleration to the environment state, which are shown in Fig.~\ref{fig:D}. Note that, the acceleration command dynamics for all related cases is for gas-engine vehicles with time constant $\tau = 0.5$s.

In Fig.~\ref{fig:K}, when trained using a point-mass kinematic model, the DRL agent achieves a near-optimal solution as compared with DP results, see the blue solid and black dashed lines. When this DRL controller is applied to car-following control with just acceleration delay, the car-following performance is degraded to a small extent. The gap-keeping error $e$ is able to return to near-zero in the steady state, see column (a) of Fig.~\ref{fig:K}. When this DRL controller is applied to car-following control with just acceleration command dynamics, the car-following performance is degraded to a bigger extent as compared to the delay case. The gap-keeping error $e$ returns to near-zero in the steady state in a longer time, see column (b) of Fig.~\ref{fig:K}. When this DRL controller is applied to car-following control with both acceleration delay and command dynamics, the performance is the worst. Both the transient and steady-state performances are significantly degraded. The steady-state error $e$ does not return to zero and forms a wavy oscillation pattern with the maximum being 0.73m and the minimum being -0.22m.

The columns (a), (b), and (c) in Fig.~\ref{fig:D} show the results for the redesigned DRL controllers trained with acceleration delay (case 2), acceleration command dynamics (case 3), and both acceleration delay and command dynamics (case 4), respectively. For all these cases, the DRL controllers achieve near-optimal solutions as compared to the DP ones. Note that the steady-state gap-keeping errors of the DP solutions in columns (b) and (c) are around 0.5 meters. This would be reduced when using a smaller interval to create the evenly spaced samples of the states for DP, although it takes much longer time to run.

\section{CONCLUSION}

By solving a particular car-following control problem using DRL (deep reinforcement learning), we show that a DRL controller trained with a point-mass kinematic model could not be generalized to solve more realistic control situations with both vehicle acceleration delay and command dynamics. We added the control inputs that are delayed and have not been executed, and the actual acceleration of the vehicle, to the reinforcement learning environment state for DRL controller development with vehicle dynamics. The training results show that this approach provides near-optimal solutions for car-following control with vehicle dynamics. In this work, the DRL controllers are trained with a fixed initial condition for all cases. We later trained a DRL controller with varying initial conditions and observed similar significant performance degradation when applying the kinematic-model-trained DRL controller to practical control with vehicle dynamics.

When the reinforcement learning environment state is augmented with the delayed control inputs, the DRL agent is expected to utilize the delayed control inputs to predict the system behavior in the future and determine the next optimal control action. Our results show that the DRL agent is capable to do so after training, in a near-optimal manner. However, because the environment state is augmented with more variables, the neural network size needs to be increased and more training time is needed, which is the disadvantage. As stated in the introduction, an alternative method is to learn the underlying dynamic system separately and use the learned system to predict the system behavior in the future after the delay time so as to determine the current control action \cite{walsh2009learning}. However, this method may not be feasible for challenging autonomous driving control systems such as merging control because such systems are subject to many variations and disturbances due to multi-vehicle interactions. It may not be easy to develop or learn an accurate model for such systems.

Future work includes developing a more robust car-following DRL controller that can be trained with rich variations of the preceding vehicle's speed. Another research direction is to develop DRL controllers with vehicle dynamics considered for more challenging autonomous driving scenarios such as freeway on-ramp merging.





\section*{ACKNOWLEDGMENT}

The authors would like to thank Toyota, Ontario Centres of Excellence, and Natural Sciences and Engineering Research Council of Canada for the support of this work.

\bibliographystyle{IEEEtran}
\bibliography{references}

\end{document}